# Testing whether linear equations are causal: A free probability theory approach

**Jakob Zscheischler**[1,2]
jakobz@tuebingen.mpg.de

**Dominik Janzing**[1]
janzing@tuebingen.mpg.de

**Kun Zhang**[1]
kzhang@tuebingen.mpg.de

1) Max Planck Institute for Intelligent Systems
Spemannstr. 38, 72076 Tübingen
Germany

2) Max Planck Institute for Biogeochemistry
Hans-Knöll-Str. 10, 07745 Jena
Germany

## Abstract

We propose a method that infers whether linear relations between two high-dimensional variables $X$ and $Y$ are due to a causal influence from $X$ to $Y$ or from $Y$ to $X$. The earlier proposed so-called *Trace Method* is extended to the regime where the dimension of the observed variables exceeds the sample size. Based on previous work, we postulate conditions that characterize a causal relation between $X$ and $Y$. Moreover, we describe a statistical test and argue that both causal directions are typically rejected if there is a common cause. A full theoretical analysis is presented for the deterministic case but our approach seems to be valid for the noisy case, too, for which we additionally present an approach based on a sparsity constraint. The discussed method yields promising results for both simulated and real world data.

## 1 Motivation

Causal analysis of high-dimensional data becomes increasingly relevant in most scientific disciplines, including, for instance, climate research, neurobiology and economy. In this paper we will significantly extend a novel bivariate causal inference method presented by Janzing et al. (2010).

Recent approaches to causal inference suggest that the joint distribution $P(X,Y)$ of two variables $X$ and $Y$ provide hints on their causal relation. In particular one can infer – with some reliability – whether $X$ causes $Y$ or $Y$ causes $X$, given that either statement is true (e.g. Friedman and Nachman (2000); Kano and Shimizu (2003); Sun et al. (2006); Hoyer et al. (2009); Daniušis et al. (2010); Mooij et al. (2009)). As opposed to conditional independence based causal discovery (Spirtes et al., 1993; Pearl, 2000), which requires more than two variables, one employs asymmetries of the joint distribution and assumes that the causal factorization $P(\text{effect}, \text{cause}) = P(\text{cause})P(\text{effect}|\text{cause})$ yields "simpler" terms than the non-causal one. None of these methods work for the case of a Gaussian joint distribution $P(X,Y)$ with a linear relation between $X$ and $Y$, because there exist a "simple" (i.e. linear) model from $X$ to $Y$ *and* vice versa. Janzing et al. (2010), however, showed that even Gaussian distributions provide hints on the causal direction if $X$ and $Y$ are high-dimensional variables. The decision whether

$$Y = AX + E \text{ with } X \perp\!\!\!\perp E \quad (1)$$

or

$$X = \tilde{A}Y + \tilde{E} \text{ with } Y \perp\!\!\!\perp \tilde{E} \quad (2)$$

is more plausible as *causal* model (given that both are valid statistical models) is based on the idea that the covariance matrix $\Sigma_X$ of the cause (given that $X$ is the cause) and the structure matrix $A$ generating the effect from the cause correspond to independent mechanisms of nature.[1] Translating this into our setting, $A$ and $\Sigma_X$ should behave like "generic" pairs of matrices and fulfill some sort of independence criterion. For instance, $n \times n$-matrices $B$ and $C$, whose entries are chosen independently satisfy the following trace condition for large $n$ with high probability:

$$\tau_n(BC) \approx \tau_n(B)\tau_n(C), \quad (3)$$

where $\tau_n(.) := \text{tr}(.)/n$ denotes the normalized trace. Significant violations of (3) are thus considered as "non-generic". Janzing et al. (2010) shows that pairs $(A, \Sigma_X)$ satisfying

$$\tau_n(A^T A \Sigma_X) \approx \tau_n(A^T A)\tau_n(\Sigma_X) \quad (4)$$

---
[1]For random variables, Lemeire and Dirkx (2006) and Janzing and Schölkopf (2010) postulate that the shortest description of $P(\text{cause}, \text{effect})$ is given by separate descriptions of $P(\text{cause})$ and $P(\text{effect}|\text{cause})$. There, description length is defined via Kolmogorov complexity, which is uncomputable.

indeed induce pairs $(\tilde{A}, \Sigma_Y)$ that violate (3) significantly. This makes the causal direction identifiable and allows for a causal inference principle that we refer to as the *Trace Method*. Janzing et al. (2010) presented encouraging experiments with real world data, but the applicability is limited by the fact that the method requires good estimates of the structure matrices $A, \tilde{A}$ and the covariance matrices $\Sigma_X, \Sigma_Y$. Hence, the number of samples needs to be significantly larger than the dimension. This is a serious limitation since, due to concentration of measure phenomena, detecting non-generic relations becomes more reliable in high dimensions.

We will extend Janzing et al. (2010) in four aspects: First, we will describe how *free probability theory* provides a general tool to formalize generic relations between high-dimensional random matrices and obtain (4) as a special case of testable conditions. Second, we will use free probability theory to show that (4) can be tested even without having good estimates for $A$ and $\Sigma_X$. Based on this insight, we will develop a method that is explicitly designed for the regime where the number of data points is smaller than the dimension. Third, we will prove that unobserved common causes also lead to violations of (4). Our method can therefore also be applied for detecting confounders confirmed by experiments on simulated data (Section 2.2). Finally, we will provide a test assessing the significance of the violation of the trace condition (Section 2.4).

**Free Probability Theory** was introduced by Voiculescu (1997) and can be seen as a generalization of classical probability theory where random variables are replaced by "non-commuting variables", which are in fact *operators*. This concept is useful for studying asymptotical relations between random matrices of increasing dimension because the operators represent, in a certain weak sense, limits of these matrices for dimension to infinity. A normalized linear form $\phi$ on the algebra of operators plays the role of the expectation $\mathbb{E}$ in classical statistics. Statistical independence between random variables $W, Z$ is replaced with *free independence* between operators $A$ and $B$.

**Definition 1 (free independence)**
Let $\mathcal{A}$ be an algebra and $\phi: \mathcal{A} \to \mathbb{R}$ a linear functional on $\mathcal{A}$ with $\phi(1) = 1$. Then $A$ and $B$ are called free if
$$\phi(p_1(A)q_1(B)p_2(A)q_2(B)\cdots) = 0$$
for polynomials $p_i, q_i$, whenever $p_i(A) = q_i(B) = 0$.

It follows that if $A$ and $B$ are free it holds $\phi(AB) = \phi(A)\phi(B)$, analog to $\mathbb{E}(WZ) = \mathbb{E}(W)\mathbb{E}(Z)$ in classical independence. One can, however, also derive $\phi(ABAB) = \phi(A^2)\phi(B)^2 + \phi(A)^2\phi(B^2) - \phi(A)^2\phi(B)^2$, which has no classical counterpart (see Lemma 1).

## 2 Theory

We will understand the postulated independence between the distribution of the cause ($\Sigma_X$) and the causal mechanism ($A$) mentioned above as free independence. Having this in mind, we choose a model that induces free independence between $\Sigma_X$ and $A^T A$ as dimensionality goes to infinity. By Lemma 1 this is obtained by fixing the structure matrix $A$ and choosing the covariance of the cause $\Sigma_X$ from a rotation invariant ensemble. This will be our model assumption throughout the paper. All limits in this section are understood in the sense of almost sure convergence. To clarify terminology, we recall that the spectral measure of an $n \times n$-matrix $D$ is given by the cumulative distribution function $F(\lambda) = \frac{|\{\lambda_D \leq \lambda\}|}{n}$, where $\lambda_D$ denotes the eigenvalues of $D$.

### 2.1 X causing Y

Although the theory is also valid if $X$ and $Y$ have different dimensions, for better comprehension we only describe the case where both are of equal dimension $n$ (i.e. the structure matrix $A$ is square). Whenever it is necessary for the discussion of the asymptotics $n \to \infty$, $n$ will occur as index of random variables and matrices. We start by defining an estimator of the matrix $A$ for the model $Y = AX + E$ with $X \perp\!\!\!\perp E$,

$$\hat{A} = \hat{\Sigma}_{YX}\hat{\Sigma}_X^+ = A\hat{\Sigma}_X\hat{\Sigma}_X^+, \tag{5}$$

where $\hat{\Sigma}$ are the standard estimators for the (cross-)covariance matrices and $(\cdot)^+$ is the Moore-Penrose pseudoinverse. This estimator is justified since it holds $\Sigma_{YX} = A\Sigma_{XX}$ for the true quantities. Note that $\hat{A}$ can be very different from the true $A$. The pseudoinverse (instead of the inverse) is used for the case that $\hat{\Sigma}_X$ does not have full rank (i.e. less samples than dimensions). To measure the violation of (4) we introduce the following scale invariant quantity:

**Definition 2 (empirical delta)**
Given observations $(x_1, y_1), \ldots, (x_k, y_k)$ with $x_i, y_i \in \mathbb{R}^n$ and $r = \text{rank}(\hat{\Sigma}_X)$, define

$$\hat{\Delta}_{X \to Y} := \log \frac{\tau_n(\hat{A}\hat{\Sigma}_X\hat{A}^T)}{\frac{n}{r}\tau_n(\hat{A}^T\hat{A})\tau_n(\hat{\Sigma}_X)}. \tag{6}$$

It is a finite sample analog to the quantity $\Delta_{X \to Y}$ in Janzing et al. (2010) and has been obtained by replacing the true covariance matrices with the empirical ones and by adding the factor $n/r$. Note that if $P(X)$ has a density and $k \leq n$ it follows $r = k - 1$ almost surely, and $r = n$ for $k > n$. The following results show that the factor $n/r$ accounts for the fact that the kernels of $\hat{A}$ and $\hat{\Sigma}_X$ coincide if the number of samples is smaller than the dimension. For the remainder of this section only the noiseless case will be discussed.

**Theorem 1 (result for forward direction)**

Assume that for a given sequence of matrices $A_n$, $\Sigma_X^{(n)}$ is chosen from a rotation invariant ensemble, i.e., $\mathcal{P}(U_n \Sigma_X^{(n)} U_n^T) = \mathcal{P}(\Sigma_X^{(n)})$ for $U_n$ uniformly drawn from the orthogonal group $O(n)$. We assume further that the spectral measures of $A_n^T A_n$ and $\Sigma_X^{(n)}$ converge weakly as $n \to \infty$.

Let $Y_n = A_n X_n$ and $c > 0$ be some constant and $k = \lceil cn \rceil$ observations drawn i.i.d. from the joint $n$-dimensional distribution $P(X_n, Y_n)$. Then

$$\lim_{n \to \infty} \hat{\Delta}_{X_n \to Y_n} = 0.$$

To prove Theorem 1 we use the following results which are based on free probability theory (Voiculescu et al., 1992; Voiculescu, 1997). The idea of the lemma below is that conjugating a matrix with a random rotation destroys dependencies to any other matrix.

**Lemma 1 (freeness by rotation invariance)**

Let $F_n, H_n$ be sequences of real symmetric deterministic matrices whose spectral measures converge weakly as $n \to \infty$. Consider

$$G_n = U_n H_n U_n^T, \qquad (7)$$

where $U_n$ is a random uniformly distributed orthogonal matrix from the ensemble $O(n)$. Then $F_n$ and $G_n$ become free in the limit $n \to \infty$ and it holds

$$\lim_{n \to \infty} \tau_n(F_n G_n) = \lim_{n \to \infty} \tau_n(F_n) \tau_n(G_n), \quad \text{and} \quad (8)$$

$$\lim_{n \to \infty} \tau_n(G_n F_n G_n F_n) = \lim_{n \to \infty} [\tau_n(F_n^2) \tau_n(G_n)^2 + \tau_n(F_n)^2 \tau_n(G_n^2) - \tau_n(F_n)^2 \tau_n(G_n)^2]. \quad (9)$$

**Sketch of proof of Lemma 1**: We briefly outline how free probability theory is applied. If all expressions $\tau_n(F_n^l)$ and $\tau_n(G_n^l)$ ($l \in \mathbb{N}$) converge for $n \to \infty$, one introduces "limit objects" $F, G$ for the matrix ensembles $F_n$ and $G_n$ and a linear functional $\phi$ satisfying $\phi(F^l) := \lim_{n \to \infty} \tau_n(F_n^l)$ (similarly for $G$). With our assumptions on $F_n$ and $G_n$, $F$ and $G$ become free (Section 9.2 in Speicher (1997), Voiculescu et al. (1992)). From the definition of freeness we get in particular $\phi(FG) = \phi(F)\phi(G)$, and $\phi(FGFG) = \phi(F^2)\phi(G)^2 + \phi(F)^2\phi(G^2) - \phi(F)^2\phi(G)^2$. Translating these equations into statements about limits of $\tau_n$ yields (8) and (9), respectively. □

**Proof of Theorem 1:** For better readability the index $n$ of the matrices is omitted. Since for given $A^T A$, $\Sigma_X$ is drawn from a rotation invariant ensemble, due to i.i.d. sampling, $\hat{\Sigma}_X \hat{\Sigma}_X^+$, too, comes from a rotation invariant ensemble. Both matrices become free in the limit due to Lemma 1. We apply (8) and the definition of $\hat{A}$ in (5). Consider first the enumerator of (6):

$$\lim_{n \to \infty} \tau_n(\hat{A} \hat{\Sigma}_X \hat{A}^T) = \lim_{n \to \infty} \tau_n(\hat{\Sigma}_{YX} \hat{\Sigma}_X^+ \hat{\Sigma}_X \hat{\Sigma}_X^+ \hat{\Sigma}_{XY})$$

$$= \lim_{n \to \infty} \tau_n(A \hat{\Sigma}_X A^T)$$

$$\stackrel{(8)}{=} \lim_{n \to \infty} \tau_n(A^T A) \tau_n(\hat{\Sigma}_X).$$

Now we look at the denominator of (6):

$$\lim_{n \to \infty} \frac{n}{r} \tau_n(\hat{A}^T \hat{A}) \tau_n(\hat{\Sigma}_X)$$

$$= \frac{n}{r} \lim_{n \to \infty} \tau_n(\hat{\Sigma}_X^+ \hat{\Sigma}_X A^T A \hat{\Sigma}_X \hat{\Sigma}_X^+) \tau_n(\hat{\Sigma}_X)$$

$$\stackrel{(8)}{=} \frac{n}{r} \lim_{n \to \infty} \tau_n(\hat{\Sigma}_X \hat{\Sigma}_X^+) \tau_n(A^T A) \tau_n(\hat{\Sigma}_X)$$

$$= \lim_{n \to \infty} \tau_n(A^T A) \tau_n(\hat{\Sigma}_X).$$

Here we used that $\hat{\Sigma}_X \hat{\Sigma}_X^+$ is a projection matrix with trace $r$. □

Next we show how the variances of the spectral measures of the estimated structure matrix and the true structure matrix are related to each other. This will be used in the proof of Theorem 2.

**Lemma 2**

Let $\hat{Z}_n$ and $Z_n$ be real valued random variables whose distribution is given by the spectral measure of the non-zero eigenvalues of $\hat{A}_n^T \hat{A}_n$ and $A_n^T A_n$, respectively. Let $Z_n$ converge in distribution to a random variable $Z$ with finite moments. Then also $\hat{Z}_n$ converges in distribution to some $\hat{Z}$ with

$$\text{var}(\hat{Z}) = \frac{r}{n} \text{var}(Z).$$

**Sketch of proof:** We introduce $Q := \lim_{n \to \infty} \hat{\Sigma}_X^{(n)} \hat{\Sigma}_X^{(n)+}$ and $\mathbb{A} := \lim_{n \to \infty} A_n^T A_n$. These limit objects are well-defined in free probability theory provided that the spectral measures converge in distribution (Voiculescu, 1997). We then consider the algebra generated by the free variables $Q$ and $\mathbb{A}$ and use $\phi$ to express the limits of $\tau_n$. It holds

$$\text{var}(\hat{Z}) = \lim_{n \to \infty} \text{var}(\hat{Z}_n)$$

$$= \lim_{n \to \infty} \left[ \frac{n}{r} \tau_n \left( (\hat{A}_n^T \hat{A}_n)^2 \right) - \left( \frac{n}{r} \tau_n(\hat{A}_n^T \hat{A}_n) \right)^2 \right]$$

$$= \frac{n}{r} \phi \left( (Q \mathbb{A} Q)^2 \right) - \left( \frac{n}{r} \phi(Q \mathbb{A} Q) \right)^2$$

$$\stackrel{(8)}{=} \frac{n}{r} \phi(\mathbb{A} Q \mathbb{A} Q) - \left( \frac{n}{r} \phi(Q) \phi(\mathbb{A}) \right)^2 \qquad (10)$$

$$\stackrel{(9)}{=} \frac{n}{r} \left( \frac{r^2}{n^2} \phi(\mathbb{A}^2) + \frac{r}{n} \phi(\mathbb{A})^2 - \frac{r^2}{n^2} \phi(\mathbb{A})^2 \right) - \phi(\mathbb{A})^2$$

$$= \frac{r}{n} \left( \phi(\mathbb{A}^2) - \phi(\mathbb{A})^2 \right) = \frac{r}{n} \text{var}(Z).$$

Use of Theorem 1 has been made in (10). □

**Theorem 2 (violation in backward direction)**
*We adopt the assumptions of Theorem 1 and Lemma 2 and assume furthermore that $\hat{Z}_n^{-1}$ converges in distribution to $\hat{Z}^{-1}$. Then,*

$$\lim_{n \to \infty} \hat{\Delta}_{Y_n \to X_n} < 0,$$

*whenever $Z$ has non-zero variance.*

**Proof:** Again the index $n$ is omitted. We denote the estimated structure matrix for the backwards direction by $\tilde{\hat{A}} = \hat{\Sigma}_{XY} \hat{\Sigma}_Y^+$. First we observe that

$$\tilde{\hat{A}} \hat{\Sigma}_Y \tilde{\hat{A}}^T = \hat{\Sigma}_X \quad \text{and} \quad \hat{\Sigma}_Y = \hat{A} \hat{\Sigma}_X \hat{A}^T.$$

Using Theorem 1 we obtain

$$\lim_{n \to \infty} \hat{\Delta}_{Y \to X} = \lim_{n \to \infty} \log \frac{\tau_n(\tilde{\hat{A}} \hat{\Sigma}_Y \tilde{\hat{A}}^T)}{\frac{n}{r} \tau_n(\tilde{\hat{A}}^T \tilde{\hat{A}}) \tau_n(\hat{\Sigma}_Y)}$$
$$= -\log \lim_{n \to \infty} \frac{n}{r} \tau_n(\tilde{\hat{A}}^T \tilde{\hat{A}}) \frac{n}{r} \tau_n(\hat{A}^T \hat{A}).$$

One can show that $(\hat{A}^T \hat{A})^+ = (\tilde{\hat{A}} \tilde{\hat{A}}^T)$. The $r$ non-zero eigenvalues of $\hat{A}^T \hat{A}$ and $\tilde{\hat{A}}^T \tilde{\hat{A}}$ are thus inverse to each other. Hence, $\frac{n}{r} \tau_n(\tilde{\hat{A}}^T \tilde{\hat{A}}) = \mathbb{E}(\hat{Z}_n^{-1}) \to \mathbb{E}(\hat{Z}^{-1})$. On the other hand, $\mathbb{E}(\hat{Z}^{-1}) \mathbb{E}(\hat{Z}) > 1$ due to Cauchy-Schwarz' inequality because $\hat{Z}$ has non-zero variance by Lemma 2. It follows

$$\lim_{n \to \infty} \hat{\Delta}_{Y_n \to X_n} = -\log \mathbb{E}(\hat{Z}^{-1}) \mathbb{E}(\hat{Z}) < 0. \quad (11)$$

□

The proof of Theorem 2 shows that the strength of the violation of the trace condition in the backward direction essentially depends on the eigenvalue distribution of $A^T A$, particularly on its variance, and the quotient $r/n$ (if either the variance is zero or $r/n \to 0$, the inequality in (11) does not hold).

### 2.2 Detecting confounders

In this section we want to discuss how the previously shown results may even help to detect a hidden common cause.

**Theorem 3 (violation by confounding)**
*Assume that for a given sequence of $A_n$, the sequences of matrices $B_n^T B_n$ and $\Sigma_Z^{(n)}$ are drawn from a prior satisfying the symmetry condition $\mathcal{P}(U \Sigma_Z^{(n)} U^T, V B_n^T B_n V^T) = \mathcal{P}(\Sigma_Z^{(n)}, B_n^T B_n)$, with $U, V \in O(n)$.*

*Let $P(X_n, Y_n)$ be generated by the latent model $X_n = A_n Z_n, Y_n = B_n Z_n$. Then it holds*

$$\lim_{n \to \infty} \hat{\Delta}_{X_n \to Y_n} < 0 \quad \text{and} \quad \lim_{n \to \infty} \hat{\Delta}_{Y_n \to X_n} < 0,$$

*whenever the spectral measures of $A_n^T A_n$ and $B_n^T B_n$ converge weakly to distributions with non-zero variance.*

**Proof:** The index $n$ for the matrices is omitted. By the definition of $\hat{\Delta}_{X_n \to Z_n}$ and Theorem 2 we obtain

$$\hat{\Delta}_{X_n \to Z_n} = \log \frac{\tau_n(\hat{\Sigma}_{ZX} \hat{\Sigma}_X^+ \hat{\Sigma}_{XZ})}{\frac{n}{r} \tau_n(\hat{\Sigma}_X^+ \hat{\Sigma}_{XZ} \hat{\Sigma}_{ZX} \hat{\Sigma}_X^+) \tau_n(\hat{\Sigma}_X)} < 0,$$

and $\hat{\Delta}_{X_n \to Y_n}$ is formally the same expression with replacing $\hat{\Sigma}_{XZ}$ and $\hat{\Sigma}_{ZX}$ with $\hat{\Sigma}_{XY}$ and $\hat{\Sigma}_{YX}$, respectively. Due to $\hat{\Sigma}_{YX} = B \hat{\Sigma}_{ZX}$ and $\hat{\Sigma}_{XY} = \hat{\Sigma}_{XZ} B^T$ we get

$$\hat{\Delta}_{X_n \to Y_n} = \log \frac{\tau_n(B \hat{\Sigma}_{ZX} \hat{\Sigma}_X^+ \hat{\Sigma}_{XZ} B^T)}{\frac{n}{r} \tau_n(\hat{\Sigma}_X^+ \hat{\Sigma}_{XZ} B^T B \hat{\Sigma}_{ZX} \hat{\Sigma}_X^+) \tau_n(\hat{\Sigma}_X)}.$$

Lemma 1 implies

$$\lim_{n \to \infty} \tau_n(BKB^T) = \lim_{n \to \infty} \tau_n(K) \tau_n(BB^T)$$

where $K$ incorporates the conjunction of the other matrices. This follows from the cyclicity of the trace and because $BB^T$ is drawn from a rotation invariant ensemble for fixed $K$. Finally, canceling out $\lim_{n \to \infty} \tau_n(BB^T)$ yields

$$\lim_{n \to \infty} \hat{\Delta}_{X_n \to Y_n} = \lim_{n \to \infty} \hat{\Delta}_{X_n \to Z_n} < 0.$$

The second inequality follows by exchanging $X$ and $Y$, and $A$ and $B$, respectively. □

If one of the structure matrices has a nearly constant spectral measure the confounder $Z$ would not be detectable since the trace condition for only one direction would be violated significantly.

At this point a short note should be made regarding the case where $X$ and $Y$ have different dimensions; let them be $n$ and $m$, respectively. Assuming that $m = \lceil dn \rceil$ for some constant $d > 0$, all statements are still true. Merely an additional constant $\log \frac{m}{n}$ enters in the proof of Theorem 2.

### 2.3 Noisy case

The non-confounded noisy case will be briefly discussed here. In the noiseless case, for $\tau_n(A^T A)$, $\tau_n(A^T A \Sigma_X)$ and $\tau_n(\Sigma_X)$ consistent estimators exist in the sense that with increasing dimension these quantities converge to their exact values. In the noisy case, however, this does not hold. For the model $Y = AX + E$ with $X \perp\!\!\!\perp E$ consider for instance

$$\tau_n(\hat{A}^T \hat{A} \hat{\Sigma}_X) = \tau_n(\hat{\Sigma}_X^+ \hat{\Sigma}_{XY} \hat{\Sigma}_{YX} \hat{\Sigma}_X^+ \hat{\Sigma}_X)$$
$$= \tau_n(A^T A \hat{\Sigma}_X) + 2 \tau_n(A \hat{\Sigma}_X \hat{\Sigma}_X^+ \hat{\Sigma}_{XE})$$
$$+ \tau_n(\hat{\Sigma}_{EX} \hat{\Sigma}_X^+ \hat{\Sigma}_{XE}).$$

$\hat{\Sigma}_{XE}$ tends to zero with increasing sample size. However, under the condition $k \leq n$, it holds

$$\tau_n(\hat{\Sigma}_{EX}\hat{\Sigma}_X^+\hat{\Sigma}_{XE}) = \tau_n(\hat{\Sigma}_E) \neq 0\,,$$

and therefore the canonical generalization of Theorem 1 does not hold. In order to still obtain identifiability, further assumptions are necessary. We outline an idea for sparse $A$ and isotropic noise.

**Dimensionality reduction.** Our aim is to find a consistent estimator of $A$ in the forward direction. In order to achieve that, some assumptions have to be made to avoid the ill-posedness of the problem. Here it is assumed that $A$ is sparse enough, i.e. there are just a small number of non-zero entries in $A$, and as a consequence, certain techniques could estimate it consistently. In addition, it is assumed that the noise covariance matrix $\Sigma_E$ is diagonal, i.e. the noise is uncorrelated across dimensions. Each row of $A$ can be estimated separately under the causal hypothesis $X \to Y$ (and $\tilde{A}$ under $Y \to X$). Hence, without loss of generality, we consider how to estimate $A_i$, the $i$th column of $A$, in the regression problem $Y_i = A_i^T X + E_i$ under the sparsity condition on $A_i$.

We are concerned with the case where $n > k$. As proposed in Fan and Lv (2008), in this very high dimensional case, $A_i$ can be estimated by a two-scale scheme: a crude large scale screening procedure followed by a moderate scale variable selection. The screening techniques, such as sure independence screening (SIS) by Fan and Lv (2008), reduce $n$ to a reasonable scale that is below $k$. This scheme is adopted here. First, insignificant variables $X_i$ are screened out by reducing $n$ to a reasonable number which is about $\frac{k}{3}$ with least angle regression (LARS, Efron et al. (2004), when $\frac{n}{k} < 10$) or SIS (when $\frac{n}{k} \geq 10$). Second, ordinary least squares is used to estimate the coefficients in $A_i$ that correspond to the predictors selected in the first stage; note that the estimate obtained in this stage is statistically consistent given that the first stage does not screen out the predictors $X_i$ with non-zero coefficients.

### 2.4 Significance assessment

A statistical test based on an idea mentioned in Janzing et al. (2010) is presented. We modified it for the small sample regime. For simplicity we discuss only the direction $X \to Y$. $\tau_n(\hat{A}^T\hat{A}\hat{\Sigma}_X)$ is used as the test statistic. The basic assumption of our method is that for a given structure matrix $A$ (and thus $A^TA$) the covariance matrix $\Sigma_X$ of the cause is chosen from a rotation invariant ensemble. This is our null hypothesis $H_0$ for which the trace condition necessarily holds. Hence, we will reject $H_0$ if the trace condition is violated. In the small sample case, $\Sigma_X$ does not have full rank. The null distribution is difficult to calculate analytically. For this reason we will generate it empirically with Monte Carlo simulations as explained below. By definition (see (5)), $\ker(\hat{A}) = \ker(\hat{\Sigma}_X)$. We first perform a truncated eigenvalue decomposition of the covariance matrix $\hat{\Sigma}_X$,

$$\hat{\Sigma}_X = VSV^T,$$

where $S$ is a diagonal matrix containing the $r$ non-zero eigenvalues of $\hat{\Sigma}_X$ and $V \in \mathbb{R}^{n \times r}$ contains the $r$ eigenvectors of $\hat{\Sigma}_X$ as columns. $V^T$ is a mapping from $\mathbb{R}^n$ onto the image space of $\hat{\Sigma}_X$ whereas $V$ is an embedding of this image space into $\mathbb{R}^n$. We draw uniformly distributed random rotations $U_r \in O(r)$ and multiply them by $V$ and $V^T$ from the left and right, respectively, i.e. $R := V U_r V^T$. We define a random variable $W : O(n) \to \mathbb{R}$ via the trace of the estimated structure matrix multiplied with random rotations of the empirical covariance matrix,

$$W(U_r) := \tau_n(\hat{A}^T\hat{A}R\hat{\Sigma}_X R^T)\,.$$

The null distribution we are looking for is given by the distribution of $W$ induced by the uniform distribution on $O(n)$. By generating random rotations we obtain an empirical distribution with samples $w$ from $W$. Let $\bar{w}$ denote the sample median. The two-sided $p$-value $p_{X \to Y}$ for the forward model is given by

$$p_{X \to Y} = \begin{cases} 2 \cdot \#\{w \leq W(I)\}/N & \text{if} \quad W(I) \leq \bar{w} \\ 2 \cdot \#\{w \geq W(I)\}/N & \text{if} \quad W(I) > \bar{w}\,. \end{cases}$$

Here, $N$ denotes the number of drawn rotations.

Below we argue that if $Y$ (!) is the cause and Theorem 2 or Theorem 3 holds, the $p$-value for the wrong direction $p_{X \to Y}$ tends to zero as $n$ goes to infinity. As usual $\frac{r}{n} \to c$. First observe that by Lemma 1 for a sample $w$ it holds

$$\begin{aligned}\lim_{n \to \infty} w &= \lim_{r \to \infty} \frac{r}{n}\tau_r(V^T\hat{A}^T\hat{A}V\ U_r V^T\hat{\Sigma}_X V U_r^T) \\ &\stackrel{(8)}{=} \lim_{r \to \infty} \frac{r}{n}\tau_r(V^T\hat{A}^T\hat{A}V)\tau_r(S) \\ &= \lim_{r \to \infty} \frac{n}{r}\tau_n(\hat{A}^T\hat{A})\tau_n(\hat{\Sigma}_X)\,,\end{aligned}$$

for random $U_r \in O(r)$. Here, we used that $\tau_n(FF^T) = \frac{r}{n}\tau_r(F^TF)$ for an $n \times r$-matrix $F$. Obviously $\bar{w}$ has the same limit. Hence $w - \bar{w} \to 0$. On the other hand, $W(I) = \tau_n(\hat{A}^T\hat{A}\hat{\Sigma}_X)$ remains smaller than $\bar{w}$ even in the limit since Theorem 2 implies

$$\lim_{n \to \infty} \tau_n(\hat{A}^T\hat{A}\hat{\Sigma}_X) < \lim_{n \to \infty} \frac{n}{r}\tau_n(\hat{A}^T\hat{A})\tau_n(\hat{\Sigma}_X)\,.$$

With the assumptions of Theorem 2 (or Theorem 3), the above derivation shows that the wrong direction always gets rejected in the limit $n \to \infty$ (i.e. the type 2 error tends to zero for growing dimension).

# 3 Algorithm and experiments

Using both Theorem 1 and Theorem 2 combined with the statistical test from Section 2.4 we propose Algorithm 1. We call it the *Trace Method*. Here, $\alpha$ is the significance level. The algorithm works for both the large and the small sample case. Experiments should clarify what dimensionality and sample size suffices to obtain a significant violation of the trace condition. We are also interested in whether our strongly idealized model assumptions are appropiate for real world data which often contain a lot of noise. We made the observation that although the significance assessment sometimes frequently rejects both directions, yet the values of delta (see (6)) differ systematically from each other. This may be a hint for a confounder that superposes the underlying causal link. For that reason, in the last real world experiment we provide a comparison between the values of delta (Section 3.3).

---
**Algorithm 1** Trace Method
1: **Input:** $(x_1, y_1), \ldots, (x_k, y_k), \alpha$
2: Compute $\hat{A} = \hat{\Sigma}_{YX} \hat{\Sigma}_X^+$ and $\hat{\tilde{A}} = \hat{\Sigma}_{XY} \hat{\Sigma}_Y^+$
3: Compute $p$-values $p_{X \to Y}$ and $p_{Y \to X}$
4: **if** $p_{X \to Y} > \alpha$ and $p_{Y \to X} < \alpha$ **then**
5:     write "$X$ is the cause"
6: **else if** $p_{Y \to X} > \alpha$ and $p_{X \to Y} < \alpha$ **then**
7:     write "$Y$ is the cause"
8: **else if** $p_{X \to Y} < \alpha$ and $p_{Y \to X} < \alpha$ **then**
9:     write "there is a confounder or the model assumptions are violated"
10: **else**
11:     write "cause cannot be identified"
12: **end if**

---

## 3.1 Simulated data

We randomly generate models $Y = AX + \sigma E$ as follows: To build the structure matrix $A$ we independently draw diagonal elements of an $n \times n$ matrix from a Gaussian distribution and multiply this matrix by random $n$-dimensional orthogonal matrices from the left and the right, respectively. To generate a random covariance matrix we similarly draw the diagonal of an $n \times n$ matrix and conjugate it with a random orthogonal matrix (we computed random orthogonal matrices following the algorithm of Stewart (1980)). The covariance $\Sigma_E$ of the noise is generated similarly. Here, however, we introduced an adjustable parameter $\sigma$ to govern the scaling of the noise with respect to the signal: $\sigma = 0$ results in the deterministic setting, while $\sigma = 1$ puts the power of the noise to that of the signal. For the deterministic setting, $p$-values for both directions that do not exceed a significance level of 0.01 are shown in Figure 1 (a). For the correct direction we expect a value around 1% whereas for the wrong direction we expect a value of 100%. Although our theory was only presented for the deterministic case, we did some experiments in the low noise regime (b). We chose $\sigma = 0.3$. The results suggest that with reasonable small noise the method is still able to identify the causal direction. We also investigated the confounded deterministic model $X = AZ, Y = BZ$. Here we chose $Z$ in the same way as we generated $X$ above. The matrices $A$ and $B$ were chosen like $A$ in the two-variable setting. (c) shows the results. Finally we tested the high-noise setting ($\sigma = 1$) with the dimensionality reduction approach and sparse $A$ (80% zeros, results are shown in (d)). In all experiments $k = \lfloor n/2 \rfloor$.

## 3.2 Semi-empirical data

**Climate data**
A typical application area with high-dimensional data is climate research. In particular, we considered so-called "reanalysis" data[2]. There, most of the variables are computed from 3D climate models with the input of some observed climate variables. One may think of these data as "semi-empirical data". The data are given as 750 monthly mean values (1/1948 until 6/2010) on $192 \times 94$ points on a regular grid over the earth surface. We investigated the following variables:

| id | variable name | abbr |
|---|---|---|
| $X_1$ | precipitation rate | (prate) |
| $X_2$ | volumetric soil moisture | (soilm) |
| $X_3$ | specific humidity at two meters | (sphum) |
| $X_4$ | clear sky downward longwave flux | (csdlf) |
| $X_5$ | air temperature at two meters | (air2m) |
| $X_6$ | net beam downward solar flux | (nbdsf) |
| $X_7$ | upward longwave radiation flux | (uwlrf) |

We tested the following variable pairs:

1: $X_1 \stackrel{?}{\leftrightarrow} X_2$,    2: $X_1 \stackrel{?}{\leftrightarrow} X_3$,
3: $X_4 \stackrel{?}{\leftrightarrow} X_5$,    4: $X_6 \stackrel{?}{\leftrightarrow} X_7$.

We only considered data points given over land surface. This gave us 5914 locations (points in space), so the data matrix is $5914 \times 750$. Different samples are given by the values of the respective variable for different months. Human reasoning suggests the following causal directions: Rain wets the earth surface and influences soil moisture and humidity, so $X_1 \to X_2$ and $X_1 \to X_3$; clear sky downward longwave flux directly affects temperature, so $X_4 \to X_5$; near infrared downward solar flux measures a component of the energy coming from the sun which partly is reflected on the earth surface. A part of this reflected energy is upward longwave radiation, so $X_6 \to X_7$.

---
[2]We thank the NCEP/NCAR 40-year reanalysis project for providing the data (Kalnay et al., 1996).

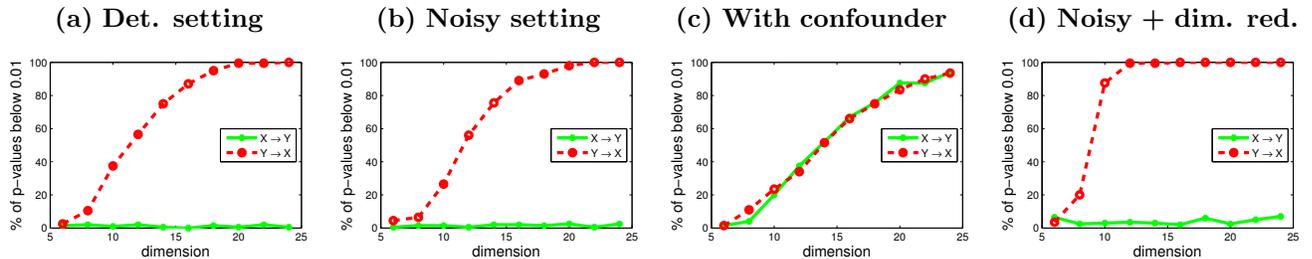

Figure 1: Percentage of $p$-values below $\alpha = 0.01$. Sample size is half of the dimension. Green solid lines correspond to the direction $X \to Y$, we expect a value around 1% except for the confounded case. Red dashed lines denote the direction $Y \to X$. (a) Deterministic setting, $X$ is cause. (b) Noisy setting with $\sigma = 0.3$, $X$ is cause. (c) Confounded setting. (d) Noisy setting using dimensionality reduction, $\sigma = 1$ and $A$ contains 80% zeros, $X$ is cause.

We provide results including significance assessment with 1000 sampled rotations. We randomly chose 300 locations, took every 11th month into account and conducted 500 experiments. This results in 300 dimensions and 68 samples. The 11th month step was chosen to weaken dependences between samples. For a significance value of 1% results are shown in Table 1.

We obtained very good results for the first and the third pair. Moreover, the wrong direction was rejected in all cases. The true directions for the second and fourth pair, however, were also quite often rejected. This may be either due to a confounder (which is more plausible for the second pair) or feedback (which is more plausible for the last pair). Also the model assumption may not be met.

| $p$-value | # val. $< \alpha$ | $p$-value | # val. $< \alpha$ |
|---|---|---|---|
| $p_{X_1 \to X_2}$ | 1 % | $p_{X_2 \to X_1}$ | 100 % |
| $p_{X_1 \to X_3}$ | 99 % | $p_{X_3 \to X_1}$ | 100 % |
| $p_{X_4 \to X_5}$ | 7 % | $p_{X_5 \to X_4}$ | 100 % |
| $p_{X_6 \to X_7}$ | 87 % | $p_{X_7 \to X_6}$ | 100 % |

Table 1: Performance of the Trace Method on climate data, $\alpha = 0.01$.

### 3.3 Real world data

**German Rhine data**
We analyzed daily values of the water levels of the Rhine[3] measured at 46 different cities in Germany from 1990 to 2008. Excluding missing values we obtained 5905 samples. We clustered the cities in two parts: 23 "upstream" cities ($X$) and 23 "downstream" cities ($Y$) which constitute the dimensions. Naturally, upstream levels causally influence downstream levels. Therefore we expect $X \to Y$. To weaken dependences between samples, we choose randomly $k$ samples out of all 5905. The statistical test mostly rejects both directions, maybe due to a confounder like rain. The $p$-values, however, are related to the deltas and so we can compare the values of delta (see 6) directly. This should reveal systematic differences. To this end we introduce a parameter $\epsilon$ and decide upon the direction whose delta is closer to zero, if the difference between both deltas is greater than $\epsilon$. Figure 2 presents the fraction of right, wrong and no decisions obtained by the deltas with $\epsilon = 0.3$ from 1000 experiments. The performance for $k < 22$ (small sample regime) is good and drops at around 22 before it increases again. The reason for the drop may be the following. Due to strong correlations across dimensions the covariance matrices are badly conditioned around dimension 22 (sample size meets rank of covariance matrix), leading to numerically instable estimations of the pseudoinverse and a random decision. Overall, although without significance assessment, the results meet the expectation quite well. Accordingly, if the significance test rejects most of the directions, considering the deltas might still provide a hint for the causal direction.

The code for the proposed method is available at http://webdav.tuebingen.mpg.de/causality/.

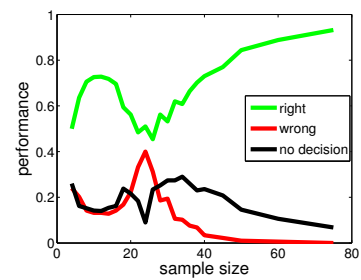

Figure 2: Performance of the Trace Method on German Rhine data. Dimensionality = 23.

---
[3] We thank the German office "Wasser- und Schiffahrtsverwaltung des Bundes" for the data.

## 4 Discussion

We presented a method that is able to distinguish between cause and effect on high-dimensional data where the sample size is smaller than the dimension. Since it is based on a trace condition we called it Trace Method. We provided the full theory for the deterministic case. Experiments on both simulated and real world data suggest that the derived algorithm may also be applicable to the non-deterministic case. Nevertheless we proposed an approach based on a sparsity constraint for this case. Furthermore, we provided a statistical test that gains reliability with increasing dimensionality. Currently, our theoretical work relies on strong and specific assumptions. Preliminary experiments, however, suggests that the method also works in a more general framework. We will work on a relaxation of the assumptions in the future.

We tested the Trace Method on simulated data to study the trade-off between sample size and number of dimensions. In fact, a high number of dimensions may replace missing samples as long as the latter still makes a significant fraction of the former (in our experiments sample sizes half of the dimension gave the best results. The method, however, still works quite well with much less samples). This rather counterintuitive fact is solved by free probability theory which relies on concentration of measure phenomena. The performance of the Trace Method on real world data is very promising. There, however, the statistical test tends to reject the correct direction too often. A solution may be to draw a larger amount of samples of the null distribution (here this is done with rotations) and then only compare $p$-values. Generating high-dimensional random orthogonal matrices is expensive, though, and the analytical distribution yet is not known. More tests on real world data are certainly necessary.

A possible application is of the Trace Method is to look for causal links between distant places on the globe (socalled "teleconnections"). This is an open problem in climate research. Different grid points can be seen as different dimensions and in such a way one may investigate the relationships between distant regions.

**Acknowledgements**

DJ has been supported by Deutsche Forschungsgemeinschaft, SPP 1395.